\documentclass[twoside,11pt]{article}

\usepackage{todonotes}
\newcommand\note[1]{\todo[inline]{#1}}
\renewcommand\note[1]{}
\usepackage{jmlr2e}

\usepackage[inline]{enumitem}
\usepackage{amssymb}
\usepackage{wrapfig}

\usepackage{tikz}
\usetikzlibrary{shapes.geometric}
\usetikzlibrary{positioning,shapes,shadows,arrows,calc,fit}
\usetikzlibrary{intersections}
\pgfdeclarelayer{background}
\pgfdeclarelayer{foreground}
\pgfsetlayers{background,main,foreground}

\newcommand{\changed}[1]{{\color{red}#1}}
\renewcommand{\changed}[1]{#1}

\usepackage{amsmath}
\usepackage{mathtools}  
\usepackage{amsfonts}

\DeclareMathOperator*{\argmin}{arg\,min}        

\newcommand{\confspace}[0]{\pmb{\Lambda}}       
\newcommand{\conf}[0]{\pmb{\lambda}}            

\newcommand{\loss}[0]{\mathcal{L}}              
\newcommand{\cost}[0]{c}                        

\newcommand{\algo}[0]{A}                        
\newcommand{\algos}[0]{\mathbf{A}}              

\newcommand{\dset}[0]{\mathcal{D}}                          
\newcommand{\dsettrain}[0]{\dset_{\text{train}}}            
\newcommand{\dsetval}[0]{\dset_{\text{val}}}                

\newcommand{\inst}[0]{i}                    
\newcommand{\insts}[0]{\mathcal{I}}         

\newcommand{\smac}{\texttt{SMAC3}}

\usepackage{lastpage}

\jmlrheading{22}{2021}{1-\pageref{LastPage}}{8/21; Revised
11/21}{12/21}{21-0888}{Marius Lindauer, Katharina Eggensperger, Matthias Feurer, Andr\'e Biedenkapp, Carolin Benjamins, Difan Deng, Ren\'e Sass and Frank Hutter}

\ShortHeadings{SMAC3: A Versatile Bayesian Optimization Package for HPO}{Lindauer, Eggensperger, Feurer, Biedenkapp, Benjamins, Deng, Sass and Hutter}
\firstpageno{1}

\begin{document}

\title{SMAC3: A Versatile Bayesian Optimization Package\\ for Hyperparameter Optimization}

\author{\name Marius Lindauer$^1$ \email lindauer@tnt.uni-hannover.de\\
       \name Katharina Eggensperger$^2$ \email eggenspk@cs.uni-freiburg.de\\
       \name Matthias Feurer$^2$ \email feurerm@cs.uni-freiburg.de\\
       \name Andr\'e Biedenkapp$^2$ \email biedenka@cs.uni-freiburg.de\\
       \name Difan Deng$^1$ \email deng@tnt.uni-hannover.de\\
        \name Carolin Benjamins$^1$ \email benjamins@tnt.uni-hannover.de\\
        \name Tim Ruhkopf$^1$ \email ruhkopf@tnt.uni-hannover.de\\
        \name Ren\'e Sass$^1$ \email sass@tnt.uni-hannover.de\\
       \name Frank Hutter$^{2,3}$ \email fh@cs.uni-freiburg.de\\
       \addr $^1$Leibniz University Hannover,
       \addr $^2$University of Freiburg,
       \addr $^3$Bosch Center for Artificial Intelligence
}

\editor{Alexandre Gramfort}

\maketitle

\begin{abstract}
Algorithm parameters, in particular hyperparameters of machine learning algorithms, can substantially impact their performance. 
To support users in determining well-performing hyperparameter configurations for their algorithms, datasets and applications at hand, 
\smac{} offers a robust and flexible framework for Bayesian Optimization, which can improve performance within a few evaluations.
It offers several facades and pre-sets for typical use cases, such as optimizing hyperparameters, solving low dimensional continuous (artificial) global optimization problems and configuring algorithms to perform well across multiple problem instances.
The \smac{} package is available under a permissive BSD-license at \url{https://github.com/automl/SMAC3}.
\end{abstract}

\begin{keywords}
  Bayesian Optimization, Hyperparameter Optimization,
  Multi-Fidelity Optimization, Automated Machine Learning, Algorithm Configuration
\end{keywords}

\section{Introduction}

It is well known that setting hyperparameter configurations of machine learning algorithms correctly is crucial to achieve top performance on a given dataset~\citep{bergstra-jmlr12a,snoek-nips12a}. Besides simple random search~\citep{bergstra-jmlr12a}, evolutionary algorithms~\citep{olson-gecco16a} and bandit approaches~\citep{li-jmlr18a}, Bayesian Optimization~(BO)~\citep{mockus-tgo78a,shahriari-ieee16a} is a commonly used approach for Hyperparameter Optimization (HPO)~\citep{feurer-bookchapter19a} because of its sample efficiency. Nevertheless, BO is fairly brittle to its own design choices~\citep{lindauer-dso19} and depending on the task at hand different BO approaches are required. 

\smac{} is a flexible open-source BO package that 
\begin{enumerate*}[label=(\roman*)]
    \item implements several BO approaches, 
    \item provides different facades, hiding unnecessary complexity and allowing easy usages, and
    \item can thus be robustly applied to different HPO tasks.
\end{enumerate*}
Its usage in successful AutoML tools, such as \texttt{auto-sklearn}~\citep{feurer-nips2015a} and the recent version of \texttt{Auto-PyTorch}~\citep{zimmer-tpami21a}, and as part of the winning solution to the latest BBO challenge~\citep{awad-arxiv20a}, demonstrates its value as a useful tool besides academic research.

\section{Different Use Cases and Modes of \smac{}}

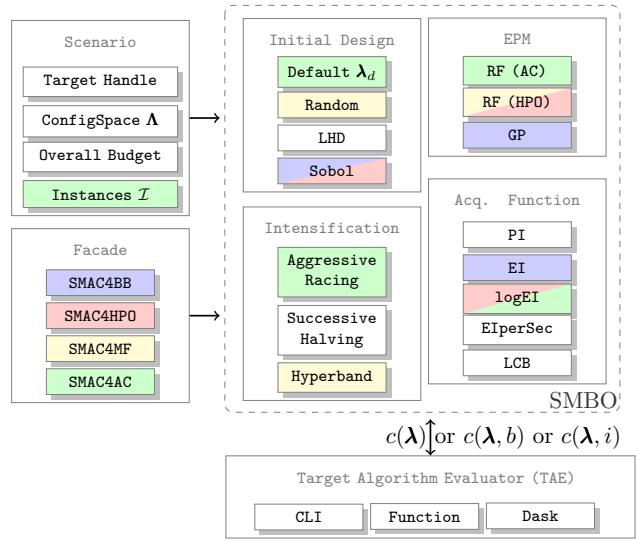
\begin{wrapfigure}[19]{r}{0.6\textwidth}
\centering
\vspace{-2em}
\scalebox{0.8}{
\tikzstyle{package}=[rectangle, draw=black, text centered, fill=white, drop shadow]
\tikzstyle{myarrow}=[->, thick]
\tikzset{
diagonal fill/.style 2 args={fill=#2, path picture={
\fill[#1, sharp corners] (path picture bounding box.south west) -|
                         (path picture bounding box.north east) -- cycle;}},
reversed diagonal fill/.style 2 args={fill=#2, path picture={
\fill[#1, sharp corners] (path picture bounding box.north west) |- 
                         (path picture bounding box.south east) -- cycle;}}
}
\begin{tikzpicture}[node distance=10em,
  title/.style={font=\scriptsize\color{black!50}\ttfamily},
  typetag/.style={rectangle, draw=black!50, font=\scriptsize\ttfamily, anchor=south}]

	\node (Scenario)[title] {Scenario};
	\node (Algo)[package, below of=Scenario, xshift=0em, typetag, node distance=2.4em, text width=6em]{Target Handle};
	\node (PCS)[package, below of=Algo, typetag, node distance=2.4em, text width=6em]{ConfigSpace $\confspace$};
	\node (Budget)[package, below of=PCS, typetag, node distance=2.2em, text width=6em]{Overall Budget};
	\node (Insts)[package, below of=Budget, typetag, node distance=2.2em, text width=6em, fill=green!20]{Instances $\insts$};
	\node (ScenBox)[draw=black!50, fit={(Scenario) (Algo) (PCS) (Insts) (Budget)}, text width=7em] {};
	
	\node (Facade)[title, below of=Scenario, node distance=9em] {Facade};
	\node (bo)[package, below of=Facade, xshift=0em, typetag, node distance=2em, text width=4em, fill=blue!20]{SMAC4BB};
	\node (hpo)[package, below of=bo, typetag, node distance=2em, text width=4em, fill=red!20]{SMAC4HPO};
	\node (bohb)[package, below of=hpo, typetag, node distance=2em, text width=4em, fill=yellow!20]{SMAC4MF};
	\node (ac)[package, below of=bohb, typetag, node distance=2em, text width=4em, fill=green!20]{SMAC4AC};
	\node (FacadeBox)[draw=black!50, fit={(Facade) (bo) (hpo) (ac) (bohb)}, text width=7em] {};
	
	\node (Init)[title, right of=Scenario, node distance=10em] {Initial Design};
	\node (first)[package, below of=Init, xshift=0em, typetag, node distance=2em, text width=4em, fill=green!20]{Default $\conf_{d}$};
	\node (random)[package, below of=first, typetag, node distance=2em, text width=4em, fill=yellow!20]{Random};
	\node (lhd)[package, below of=random, typetag, node distance=2em, text width=4em]{LHD};
	\node (Sobol)[package, below of=lhd, typetag, node distance=2em, text width=4em, diagonal fill={red!20}{blue!20}]{Sobol}; 
	\node (InitBox)[draw=black!50, fit={(Init) (Sobol) (random) (lhd) (first)}, text width=7em] {};
	
	\node (Intens)[title, below of=Init, node distance=8.06em] {Intensification};
	\node (Agr)[package, below of=Intens, xshift=0em, typetag, node distance=3em, text width=4em, fill=green!20]{Aggressive Racing};
	\node (SH)[package, below of=Agr, typetag, node distance=3.6em, text width=4em]{Successive Halving};
	\node (HB)[package, below of=SH, typetag, node distance=2.7em, text width=4em, fill=yellow!20]{Hyperband};
	\node (IntensBox)[draw=black!50, fit={(Intens) (Agr) (SH) (HB)}, text width=7em] {};
	 
	\node (EPM)[title, right of=Init, node distance=8em, yshift=0.1em] {EPM};
	\node (RF)[package, below of=EPM, xshift=0em, typetag, node distance=2em, text width=4em, fill=green!20]{RF (AC)};
	\node (RF2)[package, below of=RF, xshift=0em, typetag, node distance=2em, text width=4em, diagonal fill={red!20}{yellow!20}]{RF (HPO)}; 
	\node (GP)[package, below of=RF2, typetag, node distance=2em, text width=4em, fill=blue!20]{GP};
	\node (EPMBox)[draw=black!50, fit={(EPM) (RF) (GP)}, text width=7em] {};  
	 
	\node (Acq)[title, below of=EPMBox, node distance=4.95em] {Acq. Function};
	\node (PI)[package, below of=Acq, xshift=0em, typetag, node distance=2em, text width=4em]{PI};
	\node (EI)[package, below of=PI, typetag, node distance=2em, text width=4em, fill=blue!20]{EI};
	\node (LogEI)[package, below of=EI, typetag, node distance=2em, text width=4em, diagonal fill={green!20}{red!20}]{logEI};
	\node (EIperSec)[package, below of=LogEI, typetag, node distance=2em, text width=4em]{EIperSec};
	\node (LCB)[package, below of=EIperSec, typetag, node distance=2em, text width=4em]{LCB};
	\node (AcqBox)[draw=black!50, fit={(Acq) (PI) (EI) (LogEI) (EIperSec) (LCB)}, text width=7em] {}; 
	
	\node (TAE)[title, below of=Acq, node distance=12em, text width=13em, xshift=-3em] {Target Algorithm Evaluator (TAE)};
	\node (CMD)[package, below of=TAE, typetag, node distance=2.2em, text width=4em, xshift=-6em]{CLI};
	\node (func)[package, right of=CMD, typetag, node distance=5em, text width=4em, yshift=-0.55em]{Function};
	\node (Dask)[package, right of=func, typetag, node distance=5em, text width=4em, yshift=-0.55em]{Dask};
	\node (TAEBox)[draw=black!50, fit={(TAE) (CMD) (func) (Dask)}, text width=17em, xshift=-0em] {}; 
	
	\draw[->, thick] (ScenBox.east) to ($(ScenBox.east)+(0.5,0)$);
	\draw[->, thick] (FacadeBox.east) to ($(FacadeBox.east)+(0.5,0)$);
	
	\draw[<->, thick] (TAEBox.north) to node [right, xshift=-2.3em, text width=12em, yshift=-0em] {$c(\conf)$ or $c(\conf,b)$ or $c(\conf,\inst)$} ($(TAEBox.north)+(0,0.6)$);

	\begin{pgfonlayer}{background}
		\path (Init -| Init.west)+(-0.6,0.6) node (smacUL) {};
    	\path (LCB.east |- LCB.south)+(0.8,-0.6) node(smacBR) {};
    	\path [rounded corners, draw=black!50, dashed] (smacUL) rectangle (smacBR);
		\path (LCB.east |- LCB.south)+(0.2,-0.4) node [text=black!80] {SMBO};
    \end{pgfonlayer}
	
\end{tikzpicture}
}
\caption{Simplified overview of components in \smac{}. We color-code the different pre-set options activated by different facades.}
\label{fig:arch}
\end{wrapfigure}
\smac{} \changed{is designed to be robustly applicable to a wide range of different use-cases.
To allow for maximal flexibility, \smac{} does not only implement a Python interface, but also a CLI to communicate with arbitrary processes and programming languages.
Furthermore, \smac{} supports two kinds of parallelization techniques: (i) via DASK~\citep{rocklin-sci15a} and (ii) via running an arbitrary number of independent \smac{} instances exchanging information via the file system.
Last but not least, its modular design allows to combine different modules seamlessly. For a use-case at hand, the user only has to define a scenario, including the configuration space $\confspace$~\citep{lindauer-arxiv19x}, and choose a facade encoding different pre-sets of \smac{}, see Figure~\ref{fig:arch}. These pre-sets are specifically designed to be efficient based on the characteristics of the given use-case, as described in the following.}

\subsection{\texttt{SMAC4BB}: SMAC for Low-dimensional and Continuous Black-Box Functions}
\label{sub:bo}

The most general view on HPO is one of black-box optimization, where an unknown cost function $\cost$ is minimized with respect to its input hyperparameters $\conf \in \confspace$; one can equivalently frame this as minimizing the loss $\loss$ on validation data $\dsetval$ of a model trained on training data $\dsettrain$ with hyperparameters $\conf$:
\begin{equation}
    \conf^* \in \argmin_{\conf \in \confspace} c(\conf) = \argmin_{\conf \in \confspace} \loss(\dsettrain, \dsetval; \conf).
\end{equation}
BO with Gaussian Processes (GP) is  the traditional choice for HPO on continuous spaces with few dimensions. 
\smac{} builds on top of existing GP implementations and offers several acquisition functions, including~LCB~\citep{srninivas-icml10a}, TS~\citep{thompson-bio33a}, 
PI and EI~\citep{jones-jgo98a} and variants, e.g., EI per second~\citep{snoek-nips12a} for evaluations with different runtimes and logEI~\citep{hutter-lion10a} for heavy-tailed cost distributions. \changed{The pre-set of \texttt{SMAC4BB} follows commonly used components and comprises a Sobol sequence as initial design, a GP with 5/2-Mat\'ern Kernel and EI as acquisition function, similar to \cite{snoek-nips12a}.}

\subsection{\texttt{SMAC4HPO}: SMAC for CASH and Structured Hyperparameter Optimization}
\label{sub:cash}

\smac{} can also tackle the combined algorithm selection and hyperparameter optimization problem (CASH)~\citep{thornton-kdd13a}, and searches for a well-performing $\algo_i$ from a \emph{set of algorithms} $\algos$ and its hyperparameters $\conf \in \confspace_i$:
\begin{equation}
    (\algo^*, \conf^*) \in \argmin_{\algo_i \in \algos, \conf \in \confspace_i} c(\algo_i, \conf) = \argmin_{\algo_i \in \algos, \conf \in \confspace_i} \loss(\dsettrain, \dsetval; \algo_i(\conf)).
\end{equation}
The hyperparameter sub-space $\confspace_i$ for an algorithm $\algo_i$ is only active if $\algo_i$ was chosen. Therefore, there is a conditional hierarchy structure between the top-level hyperparameter choosing $\algo_i$ and the subspace $\confspace_i$. 
\smac{} models these spaces by using a random forest~\citep{breimann-mlj01a}, with SMAC having been the first BO approach to use this type of model~\citep{hutter-lion11a}. 
\smac{} supports multiple levels of conditionalities, e.g., one top-level hyperparameter choosing a classification algorithm and another sub-level hyperparameter 
an optimizer (for training NNs) enabling a second sub-level of hyperparameters.
\changed{The pre-set of \texttt{SMAC4HPO} is based on the tuning by \cite{lindauer-dso19} and combines a Sobol sequence as initial design, a RF as surrogate model and logEI as acquisition function.}

\subsection{\texttt{SMAC4MF}: SMAC for Expensive Tasks and Automated Deep Learning}
\label{sub:mf}

On some HPO tasks, e.g., optimizing hyperparameters and architectures for deep learning, training many models might be too expensive. 
Multi-fidelity optimization is a common approach, for cases where we can observe cheaper approximations of the true cost:
\begin{equation}
    \conf^* \in \argmin_{\conf \in \confspace} c(\conf,b_{max}) = \argmin_{\conf \in \confspace} \loss(\dsettrain, \dsetval; \conf, b_{max}).
\end{equation}
Here, a configuration is evaluated with a budget $b \le b_{max}$ (e.g., number of epochs, dataset size or channels of CNN) to obtain a cheap proxy of the true cost function at $b_{max}$. \smac{} follows the principle of BOHB~\citep{falkner-icml18a} in combining Hyperband~\citep{li-jmlr18a} and BO, where the surrogate model is fitted on the highest budget-level with sufficient observations. However, its RF models tend to yield much better performance than BOHB's TPE models, see Figure \ref{fig:comparion} for  illustrative examples. 
\changed{With the exception of the RF, the pre-set of \texttt{SMAC4MF} is similar to \cite{falkner-icml18a} and consists of a random initial design and Hyperband as intensification method.}

\subsection{\texttt{SMAC4AC}: SMAC for Algorithm Configuration}
\label{sub:ac}

A more general view on the problem of HPO is called algorithm configuration (AC)~\citep{hutter-jair09a,ansotegui-cp09a,lopez-ibanez-orp16}, where the goal is to determine a well-performing robust configuration across a set of problem instances $\inst$ from a finite set~$\insts$:
\begin{equation}
    \conf^* \in \argmin_{\conf \in \confspace} c(\conf) = \argmin_{\conf \in \confspace} \sum_{\inst \in \insts} c'(\conf, \inst)
\end{equation}
The origin of SMAC lies in AC~\citep{hutter-lion11a}.
It inherits the ideas of aggressive racing~\citep{hutter-jair09a} which  evaluates less promising candidate configurations on only a few instances, and to collect sufficient empirical evidence on many instances if the candidate could become the next incumbent configuration. Furthermore, \smac{} supports imputation of right-censored observations and a logEI to model heavy-tailed cost distributions;
it also uses a customized hyperparameter configuration of the RF surrogate model for AC. SMAC successfully optimized configuration spaces with more than $300$ hyperparameters for SAT solvers~\citep{hutter-aij17a}, yielding speedups of up to several orders of magnitude.
\changed{The pre-set of \texttt{SMAC4AC} follows \cite{hutter-lion11a} and combines a single default configuration as initial design, a RF as surrogate model (with different hyperparameter settings compared to \texttt{SMAC4HPO}), logEI as acquisition function and aggressive racing as intensification mechanism.}

\section{Brief Empirical Comparison}

\begin{figure}[tb]
    \includegraphics[width=0.32\textwidth]{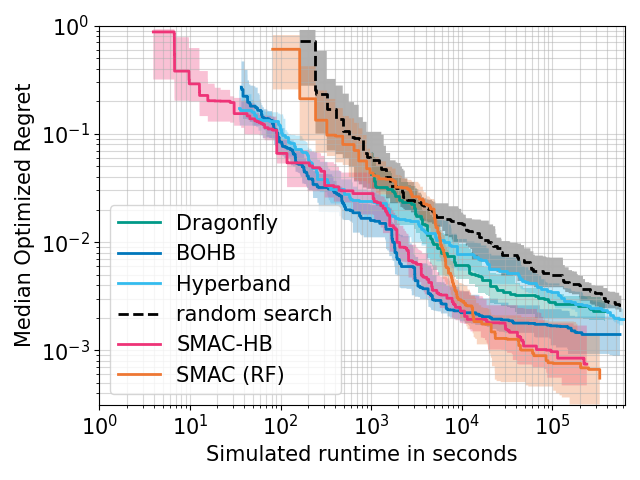}
    \includegraphics[width=0.32\textwidth]{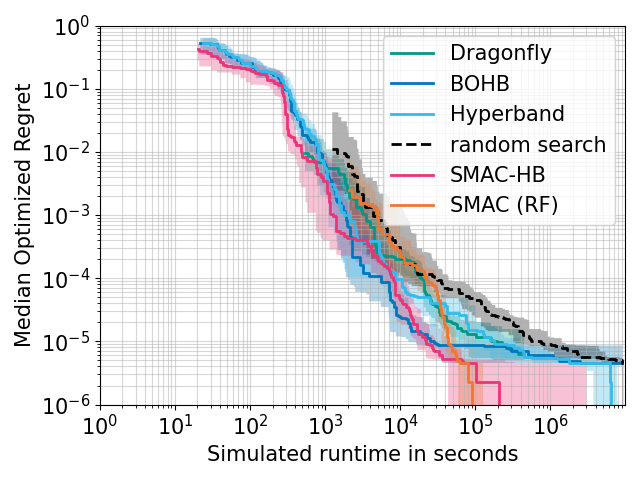}
    \includegraphics[width=0.32\textwidth]{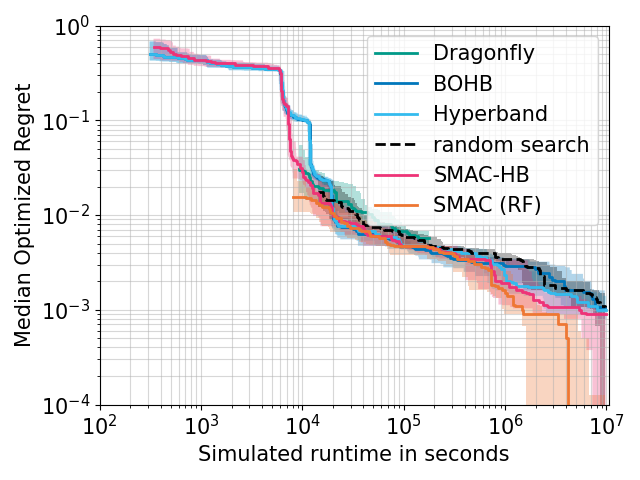}
  \caption{Comparison on Net$_{Letter}$ \changed{(6D)}, NBHPO$_{Naval}$ \changed{(9D)} and Nas1Shot1$_2$ \changed{(9D)}. \changed{Since these are all tabular or surrogate benchmarks from HPOBench, runtime is simulated by table look-ups or surrogate predictions.}}
  \label{fig:comparion}
\end{figure}

To provide an impression on the \changed{sequential} performance of \smac{}, we compared it against random search \citep{bergstra-jmlr12a}---although we consider it a weak baseline~\citep{turner-neuripscdt20a}---Hyperband~\citep{li-jmlr18a}, Dragonfly~\citep{kandasamy-jmlr20a} and BOHB \citep{falkner-icml18a} on \changed{the surrogate benchmark for HPO on DNNs on the letter dataset \citep{falkner-icml18a}}, a joint HPO+NAS benchmark~\citep{klein-arxiv19a} on the NavalPropulsion dataset and a pure NAS benchmark~\citep{zela-iclr20a}.\footnote{See  \url{https://github.com/automl/HPOBench/} for details regarding the experimental setup.} As shown in Figure~\ref{fig:comparion}, \smac{}'s multi-fidelity approach (see Sec.~\ref{sub:mf}) performs as well as Hyperband in the beginning, performs best in the middle, until \smac{}'s pure BO with RFs catches up in the end. For the whole time, \smac{} consistently outperforms Dragonfly and in the later phases, also BOHB.
\changed{For a larger empirical study, incl. \smac{}, we refer to \cite{eggensperger-arxiv21a}.}

\section{Related Work}

With the initial success of BO for HPO~\citep{hutter-lion11a,snoek-nips12a,bergstra-sci13a}, there were many follow up tools in recent years~\changed{\citep{bakshy2018ae,nardi-mascots19a,kandasamy-jmlr20a,balandat-neurips20a,li-kdd21a,scikit-opt}}. \changed{\smac{}'s advantage lies on the one hand in the efficient use of random forests as surrogate model for higher dimensional and complex spaces, and on the other hand, in its flexibility of combining different state-of-the-art BO and intensification strategies, such as aggressive racing and multi-fidelity approaches.} In addition, evolutionary algorithms are also known as efficient black-box optimizers~\citep{fortin-jmlr12a,olson-gecco16a,loshchilov-iclrws16a,nevergrad,awad-ijcai21}. Although there is the common belief that BO is particularly efficient for small budgets and evolutionary algorithms for cheap function evaluations~\citep{feurer-bookchapter19a}, choosing the right optimizer for a given task is still an open problem. To address this, first systems have been introduced that schedule several optimizers \changed{sequentially to make use of their respective strengths}~\citep{awad-arxiv20a,lan-arxiv20a,turner-neuripscdt20a}.

\section{Outlook}

Although \smac{} performs robustly on many HPO tasks, it does not exploit their landscape structure. As a next step, we plan to integrate local BO approaches~\citep{eriksson-neurips19a}. Furthermore, \smac{} provides an easy-to-use \texttt{fmin}-API and facades, but choosing \smac{}'s own hyperparameters might still be challenging~\citep{lindauer-dso19}. Therefore, we plan to add mechanisms to adaptively select \smac{}'s settings on the fly, e.g., via Bandits~\citep{hoffman-uai11a} or reinforcement learning~\citep{biedenkapp-ecai20}.

\section*{Acknowledgements}
Robert Bosch GmbH is acknowledged for financial support. This work has partly been supported by the European Research Council (ERC) under the European Union’s Horizon 2020 research and innovation programme under grant no.\ 716721. The authors acknowledge support by the High Performance and Cloud Computing Group at the Zentrum f\"{u}r Datenverarbeitung of the University of T\"{u}bingen, the state of Baden-W\"{u}rttemberg through bwHPC and the German Research Foundation (DFG) through grant no INST 37/935-1 FUGG.  

\vskip 0.2in
\bibliography{strings,lib,local,proc}

\end{document}